\documentclass[10pt,twocolumn]{article} 
\usepackage{simpleConference}
\usepackage{times}
\usepackage{graphicx}
\usepackage{amssymb}
\usepackage{url,hyperref}
\usepackage[font=normalsize, labelfont=bf]{caption}
\usepackage{amsmath}
\PassOptionsToPackage{hyphens}{url}
\usepackage{hyperref}
\usepackage[numbers,square]{natbib}

\begin{document}

\title{An Ensemble Embedding Approach for Improving Semantic Caching Performance in LLM-based Systems}

\author{
\begin{tabular}[t]{c@{\hspace{1.5em}}c@{\hspace{1.5em}}c}
Shervin Ghaffari & Zohre Bahranifard & Mohammad Akbari \\
\textit{Amirkabir Univ. of Tech.} & \textit{Amirkabir Univ. of Tech.} & \textit{Amirkabir Univ. of Tech.} \\
\textit{Dept. of Math. \& Comp. Sci.} & \textit{Dept. of Math. \& Comp. Sci.} & \textit{Dept. of Math. \& Comp. Sci.} \\
\textit{Tehran, Iran} & \textit{Tehran, Iran} & \textit{Tehran, Iran} \\
\texttt{\small shervinghaffari79@gmail.com} & \texttt{\small zohrebahraanifard@gmail.com} & \texttt{\small akbari.ma@gmail.com}
\end{tabular}
}

\maketitle

\thispagestyle{empty}

\begin{abstract}
Semantic caching enhances the efficiency of large language model (LLM) systems by identifying semantically similar queries, storing responses once, and serving them for subsequent equivalent requests. However, existing semantic caching frameworks rely on single embedding models for query representation, which limits their ability to capture the diverse semantic relationships present in real-world query distributions. This paper presents an ensemble embedding approach that combines multiple embedding models through a trained meta-encoder to improve semantic similarity detection in LLM caching systems. We evaluate our method using the Quora Question Pairs (QQP) dataset, measuring cache hit ratios, cache miss ratios, token savings, and response times. Our ensemble approach achieves a 92\% cache hit ratio for semantically equivalent queries while maintaining an 85\% accuracy in correctly rejecting non-equivalent queries as cache misses. These results demonstrate that ensemble embedding methods significantly outperform single-model approaches in distinguishing between semantically similar and dissimilar queries, leading to more effective caching performance and reduced computational overhead in LLM-based systems.
\end{abstract}

\section{Introduction}
The rapid advancement of large language models (LLMs) has transformed natural language processing applications, with models from OpenAI, Meta, and Anthropic demonstrating remarkable capabilities in text summarization, translation, content generation, and sentiment analysis \citep{brown2020language}. However, the computational expense of LLM inference, coupled with token-based pricing models, creates significant cost and latency challenges for production systems \citep{huang2024survey}.

Research indicates that approximately 31\% of ChatGPT queries exhibit semantic similarity to previously submitted requests \citep{wang2024meancache}, revealing substantial inefficiencies in current LLM deployment strategies. Since users incur costs based on both input and output tokens processed, optimizing the handling of repetitive queries presents a critical opportunity for reducing operational expenses and improving system responsiveness.

Semantic caching addresses this challenge by storing and reusing responses based on query meaning rather than exact textual matches. Unlike traditional keyword-based caching, semantic caching leverages embedding models \citep{reimers2019sentence, deerwester1990indexing, tu2024enhancing} to identify semantically equivalent queries, enabling response reuse across paraphrased or reformulated requests. This approach has been implemented in systems like GPTCache \citep{bang2023gptcache}, an open-source framework that provides modular components for embedding generation, similarity evaluation, and cache management.

Despite these advances, current semantic caching systems face a fundamental limitation: their reliance on single embedding models for query representation. Individual embedding models, while powerful, exhibit varying strengths across different semantic dimensions and query types. This limitation becomes particularly pronounced in diverse application contexts where queries span multiple domains or require different types of semantic understanding.

To address this limitation, we propose an ensemble embedding approach that combines multiple embedding models to improve semantic similarity detection in LLM caching systems.
Our method leverages the complementary strengths of different embedding architectures by first selecting base models with low-correlated embedding spaces to maximize representational diversity. We then explore both non-parametric fusion techniques and a trainable meta-encoder that learns to optimally combine these diverse representations \citep{kiela2020sentence, yin2016learning}.

We evaluate our approach using the Quora Question Pairs (QQP) dataset \citep{iyer2017quora}, measuring cache hit ratios, cache miss ratios, response time improvements, and token savings. Our ensemble method achieves over 92\% cache hit ratios for semantically equivalent queries while correctly identifying 85\% of non-equivalent queries as cache misses. These improvements translate directly to reduced response latency through increased cache utilization and decreased computational overhead through reduced LLM inference requirements.

The primary contributions of this work are: (1) a novel ensemble embedding framework for semantic caching that outperforms single-model approaches, (2) systematic evaluation of fusion techniques for combining diverse embedding representations, and (3) empirical demonstration of significant performance improvements in cache hit accuracy and system efficiency metrics.

\section{Related Work}
\label{sec:related_work}

\subsection{Semantic Caching for LLMs}
The challenge of optimizing LLM inference costs has led to significant research in semantic caching approaches. Early frameworks like GPTCache established the foundation for semantic similarity-based response reuse, utilizing single embedding models to identify semantically equivalent queries. However, recent work has identified several limitations in these baseline approaches, leading to more sophisticated caching strategies \citep{tu2024enhancing, sun2024semantic}.

SCALM (Semantic Caching for Automated LLMs) addresses inefficiencies in traditional semantic caching by incorporating pattern detection and frequency analysis \citep{li2024scalm}. The system identifies the most frequent cache entries and query patterns in LLM-based chat services, ranking queries based on detected patterns to optimize cache hit ratios. Experimental evaluation demonstrates that SCALM achieves a 63\% relative increase in cache hit ratio and a 77\% reduction in token usage compared to GPTCache. Despite these improvements, SCALM maintains the single embedding model paradigm, which limits its ability to generalize across diverse query semantics and complex real-world interactions.

Generative Caching for LLMs introduces a dynamic approach that balances cost, latency, and response quality through adaptive similarity thresholds \citep{iyengar2025generative}. The system adjusts thresholds based on query complexity, user preferences, and computational constraints. For instance, code generation queries receive higher similarity thresholds to ensure response accuracy, while simple question-answering tasks operate with lower thresholds to maximize cache utilization. The framework additionally supports response synthesis for novel queries by combining cached responses. Performance evaluation shows nine-fold speed improvements over GPTCache, with customizable parameters allowing users to select domain-appropriate embedding models.

\subsection{Ensemble Methods in Embedding Systems}
While semantic caching research has primarily focused on single embedding models, ensemble approaches have demonstrated success in related domains. In information retrieval, combining multiple embedding models has shown improvements in capturing diverse semantic relationships and handling domain-specific queries \citep{kiela2020sentence, akgul2024enhancing, bryan2024comparing}. However, these techniques have not been systematically applied to semantic caching systems, where the trade-offs between representation diversity and computational efficiency present unique challenges.

\subsection{Our Contribution}
The approaches discussed above represent significant advances in semantic caching, yet they share a common limitation: reliance on single embedding models for query representation. While Generative Caching allows model selection based on query domains, it does not leverage multiple models simultaneously to capture complementary semantic information. Our proposed ensemble embedding approach addresses this gap by combining multiple specialized embedding models through a trained meta-encoder.

Unlike existing work that focuses on threshold optimization or pattern detection, our method tackles the fundamental representation learning problem in semantic caching. By utilizing models with low-correlated embedding spaces and learning optimal fusion through a meta-encoder, we achieve improved semantic discrimination across diverse query types. This approach complements existing optimizations in threshold tuning and pattern recognition, providing a foundation for more accurate semantic similarity detection in LLM caching systems.

\section{Dataset}
\label{sec:dataset}

We selected the Quora Question Pairs (QQP) dataset for our evaluation due to its alignment with semantic caching requirements and real-world query characteristics. The dataset consists of over 400,000 question pairs collected from the Quora platform, where each pair contains two independently formulated questions submitted by different users. This structure directly mirrors the semantic caching scenario, where systems must determine whether incoming queries are semantically equivalent to previously cached queries.

Each question pair in QQP is annotated with a binary label indicating semantic equivalence: label 1 denotes semantically equivalent questions (duplicates that should result in cache hits), while label 0 indicates non-equivalent questions (unique queries requiring new LLM inference). This binary classification task precisely matches the core decision-making process in semantic caching systems, making QQP an ideal evaluation benchmark.
\section{Methodology}
\label{sec:methodology}

This section presents our ensemble embedding approach for improving semantic caching performance in LLM-based systems. We begin with an overview of the baseline semantic caching architecture, then detail our ensemble design, meta-encoder architecture, training procedures, and evaluation setup.

\subsection{Semantic Caching System Architecture}
\label{subsec:semantic_caching_architecture}

\begin{figure}[h]
\centering
\includegraphics[width=\linewidth]{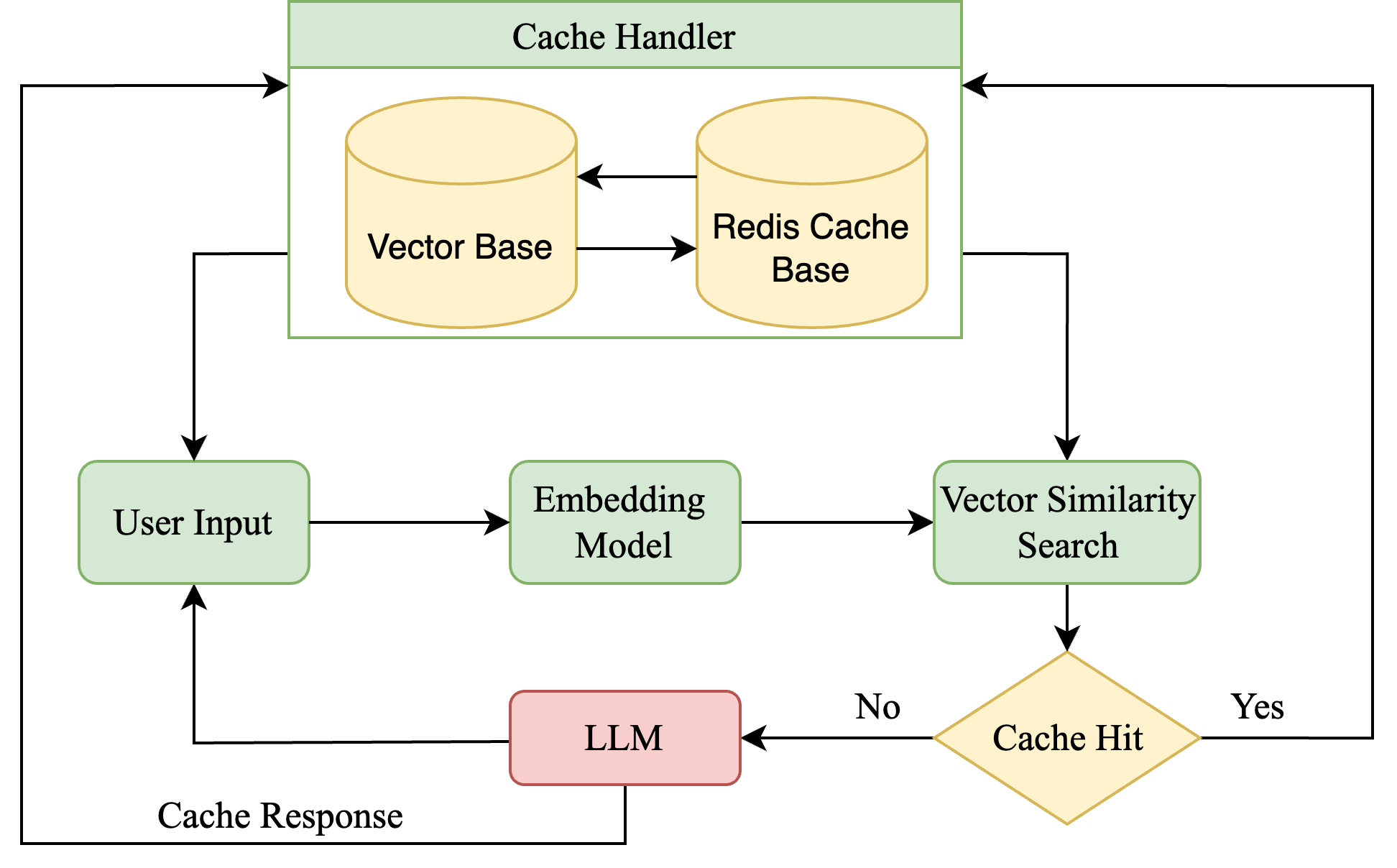}
\caption{Semantic caching system architecture showing the cache components (Cachebase and Vectorbase), and the query processing flow.}
\label{fig:system_overview}
\end{figure}

As illustrated in Figure~\ref{fig:system_overview}, when a user submits a query, the system transforms it into a numerical vector via a pre-trained embedding model. The vector’s dimension depends on the chosen model, allowing the system to compare queries and identify cached responses for similar inputs. The cache management system comprises two primary components: Cachebase stores original queries and their corresponding LLM-generated responses, while Vectorbase maintains query embeddings for semantic similarity searches. We employ Facebook AI Similarity Search (FAISS) \citep{douze2024faiss} with a Flat index for exact similarity matching, returning the top matching vectors with their similarity scores.

For cache hits (similarity scores above threshold), the system returns the cached response associated with the most similar query. For cache misses, the LLM generates a new response, which is then stored for future retrieval. Our enhancement integrates multiple embedding models through a trainable meta-encoder, processing queries through multiple models simultaneously and combining their outputs for improved similarity-based retrieval.

\subsection{Ensemble Design}
\label{subsec:ensemble_design}

Our ensemble embedding approach consists of two main components: multiple base models and a meta-model. Base models are pre-trained sentence transformers specialized in capturing different aspects of language understanding, including linguistic variations, paraphrasing, and semantic relationships. The meta-model, implemented as a trainable encoder, learns to optimally combine outputs from base models into unified representations.

The key insight behind our ensemble approach is that different embedding models capture complementary semantic information. By selecting base models with low-correlated embedding spaces and training a meta-encoder to fuse their outputs, we achieve more robust semantic representations than single-model approaches.

\subsection{Meta-Encoder Architecture}
\label{subsec:meta_encoder}

Integrating embedding representations from base models is just as critical as selecting the appropriate base models themselves. We propose an encoder-based meta-model that learns to combine embeddings dynamically through supervised training. This approach selectively amplifies the most relevant semantic signals from each base model, adapts to variations in input domain, and produces more task-aligned representations.

\subsubsection{Encoder Architecture}
\label{subsubsec:trainable_encoder_architecture}

The encoder consists of a multi-layer neural network architecture designed to process concatenated embeddings from multiple base models. The encoder features three hidden layers followed by a final projection that yields unit-length embeddings:

\textbf{Layer 1 – Initial Transformation}: Transforms the input (768 dimensions) into a higher-dimensional space 1024 via a linear layer, batch normalization, LeakyReLU, and dropout, promoting non-linearity and regularization.

\textbf{Layer 2 – Residual Connection}: Applies a linear transformation with batch normalization, LeakyReLU, and dropout, while adding the output from Layer 1 to enhance gradient flow and preserve important features.

\textbf{Layer 3 – Dimensionality Reduction}: Reduces the hidden dimension by half (from 1024 to 512 by default) using a similar operation sequence.

\textbf{Final Projection and Normalization}: Maps the output to the target dimension (384 by default) and L2-normalizes the vector to ensure effective similarity comparisons.

The dropout rate in all layers was 0.1, avoiding overfitting while maintaining essential information.

\begin{figure*}[htbp]
\centering
\includegraphics[width=\textwidth]{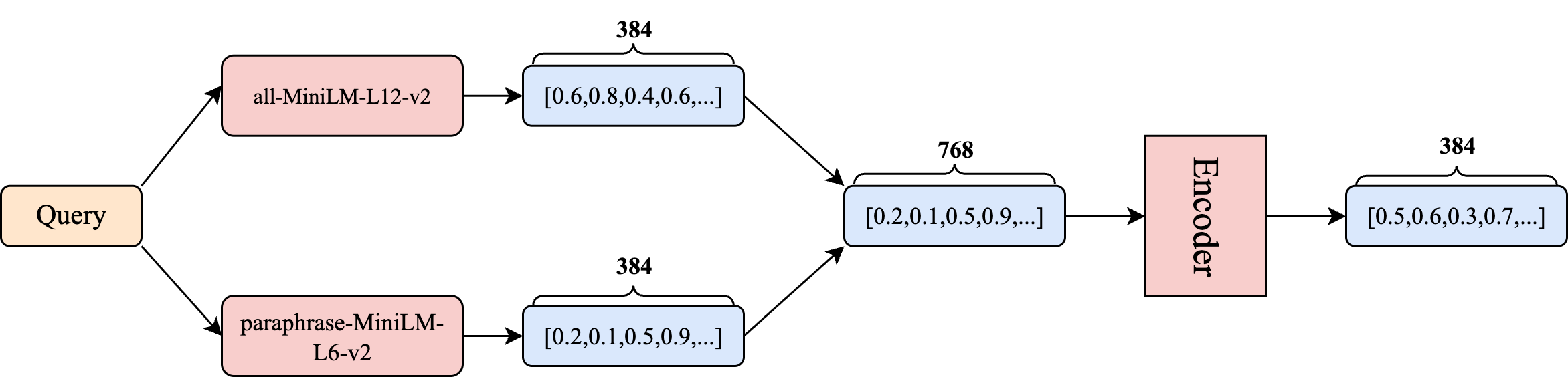}
\caption{Architecture of the proposed trainable encoder-based meta-encoder. The model dynamically integrates concatenated embeddings from multiple base models through a multi-layer neural network featuring residual connections, dimensionality reduction, and L2 normalization to produce task-aligned semantic representations.}
\label{fig:meta_learning_paradigms_comparison}
\end{figure*}

\subsubsection{Loss Function}
\label{subsubsec:loss_function}

The model is trained using a Contrastive loss function \citep{gao2021simcse} that minimizes distance between similar pairs and maximizes it for dissimilar pairs, defined in terms of cosine similarity. For a pair of embeddings, the cosine similarity is converted into a distance measure (1 – cosine similarity). The loss function penalizes duplicate pairs with high distances (squared) and non-duplicate pairs when the distance is below a predefined margin \citep{zhao2023consistent, sparck1972statistical}.

\subsection{Training Procedure}
\label{subsec:training}
The meta-encoder is trained using supervised contrastive learning \citep{gao2021simcse} to distinguish between duplicate and non-duplicate question pairs from the QQP dataset \citep{zhang2024textcg, wang2024word}.

We perform a stratified split of the dataset into training (70\%), validation (15\%), and test (15\%) subsets to maintain class balance across all partitions. Each question is embedded using the two selected pre-trained sentence embedding models, such as \textbf{all-MiniLM-L12-v2} and \textbf{paraphrase-MiniLM-L6-v2} (which will be explained in Section~\ref{subsubsec:base_model_selection_setup}), and these embeddings are concatenated and L2-normalized to form the input representations. For each pair, the encoder processes the embeddings through a shared architecture consisting of three residual blocks with batch normalization, leaky ReLU activations, dropout regularization, and a final projection layer. The outputs are again normalized to lie on the unit hypersphere.

The training process employs the following configuration:
\begin{itemize}
  \item \textbf{Optimizer:} Adam with initial learning rate of $1 \times 10^{-4}$ and weight decay of $1 \times 10^{-5}$
  \item \textbf{Learning Rate Scheduler:} Step scheduler reducing the learning rate by 0.5 every five epochs
  \item \textbf{Early Stopping:} Patience of three epochs based on validation loss
  \item \textbf{Regularization:} Dropout rate of 0.1 across all layers
\end{itemize}

Model checkpoints corresponding to the lowest validation loss are saved for downstream evaluation, and final performance is assessed on the held-out test set.

\subsection{Evaluation Setup}
\label{subsec:evaluation_setup}

This subsection details the similarity search methodology and hyperparameter optimization procedures used to evaluate our ensemble embedding approach.

\subsubsection{Similarity Search Configuration}
\label{subsubsec:similarity_search_config}

We employ cosine similarity as our distance metric and utilize FAISS \citep{douze2024faiss} for efficient similarity search. The Flat index method is selected to ensure exact nearest neighbor retrieval based on cosine similarity, trading computational cost for accuracy in similarity-based cache performance \citep{robertson2009probabilistic, mai2024improving}.

\subsubsection{Similarity Threshold Optimization}
\label{subsubsec:hyperparameter_optimization}

To classify question pairs as duplicates or non-duplicates, we need to optimize a similarity threshold based on cosine similarity scores produced by the encoder. The threshold determines the cutoff above which a pair is considered similar.

Based on validation set performance, we select an optimal threshold of 0.80 by maximizing the F1 score. This value strikes a balance between precision and recall, reducing both false positives (dissimilar pairs incorrectly classified as duplicates) and false negatives (missed duplicates). This balance is critical for downstream tasks such as information retrieval and duplicate question detection.

\begin{figure}[h]
\centering
\includegraphics[width=\linewidth]{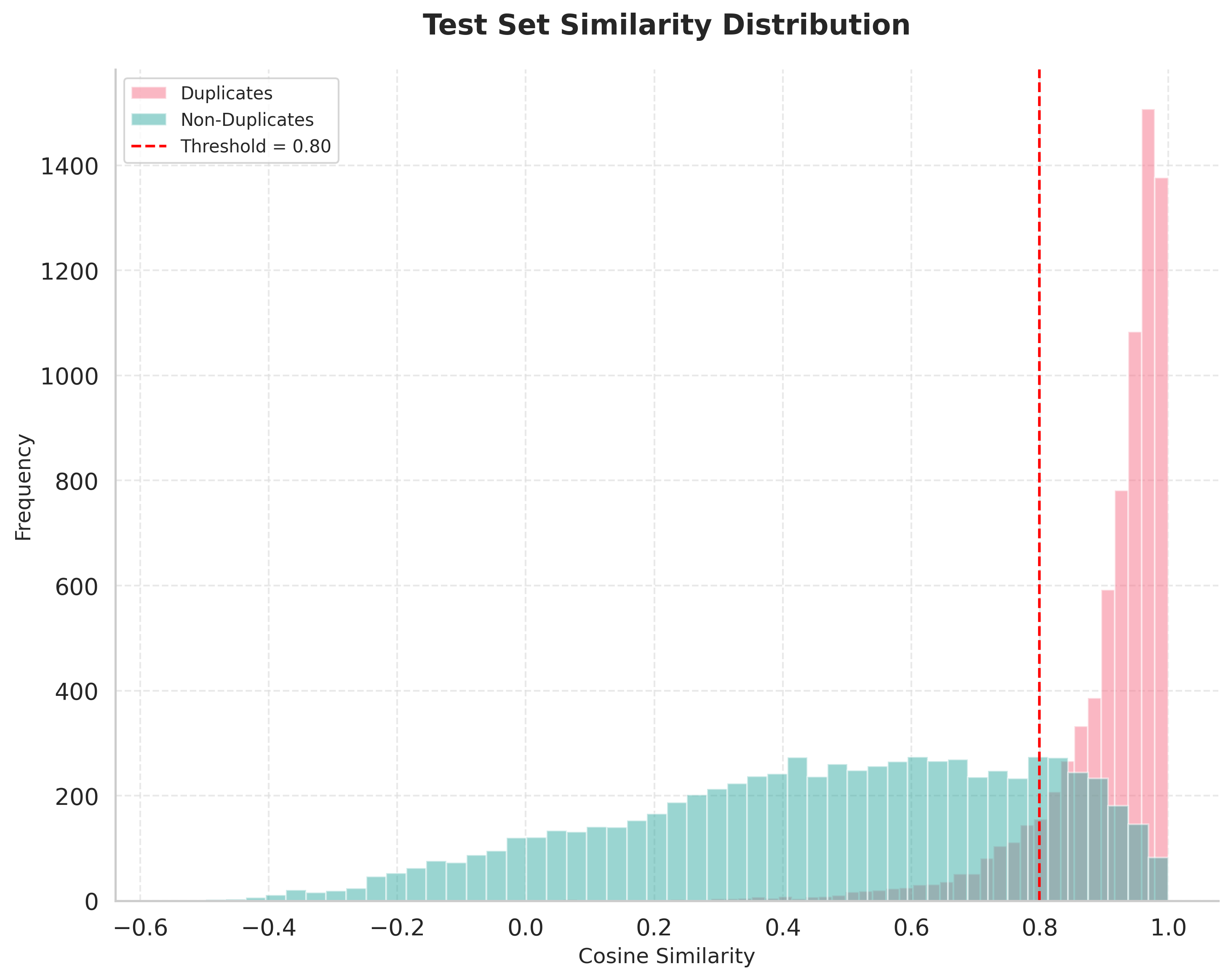}
\caption{Cosine similarity distribution on the test set. The vertical dashed line indicates the selected threshold of 0.80.}
\label{fig:similarity-distribution}
\end{figure}

Figure~\ref{fig:similarity-distribution} illustrates the distribution of cosine similarity scores between the embedding representation of paired samples for both duplicate and non-duplicate in the test set. The selected threshold separates the two classes with minimal overlap. The effectiveness of this threshold is further supported by performance metrics including precision, recall, accuracy, and F1 score, providing a robust evaluation of model behavior.

\subsubsection{Evaluation Metrics}
\label{subsubsec:evaluation_metrics}

When evaluating a caching system—particularly one based on semantic similarity—several core metrics determine its effectiveness and efficiency. Our study focuses on the hit ratio, token savings, and response time. These metrics were computed for all individual and combined embedding models using a similarity threshold of 0.8 across five different dataset sizes.

\textbf{Cache Hit Ratio:} The hit ratio reflects the percentage of requests successfully matched with semantically similar queries in the cache. This reduces redundant computation by retrieving relevant cached responses. Its complement—the miss ratio—corresponds to cases where queries are correctly identified as dissimilar and not served from the cache.

\begin{equation}
\mathrm{Hit\ Ratio} = 
\frac{
  \mathrm{Number\ of\ Cache\ Hits}
}{
  \mathrm{Total\ Number\ of\ Requests}
} \times 100
\end{equation}

\textbf{Token Saving Ratio:} Token savings quantify how much computation is avoided by serving cached responses instead of generating new ones. This includes both prompt and completion tokens.

\begin{equation}
\mathrm{Token\ Saving\ Ratio} = 
\frac{
  \mathrm{Tokens\ Served\ by\ Cache}
}{
  \mathrm{Total\ Tokens\ Processed}
} \times 100
\end{equation}

\textbf{Response Time:} This metric evaluates the average time required to serve a request. Lower response times indicate faster, more efficient caching performance.

\begin{equation}
\mathrm{Average\ Response\ Time} = 
\frac{
  \mathrm{Total\ Response\ Time}
}{
  \mathrm{Total\ Number\ of\ Requests}
}
\end{equation}

\section{Experiment Setup}
\label{sec:experiments}
We evaluate our ensemble embedding approach through comprehensive experiments comparing it against individual embedding models and baseline fusion techniques. Our evaluation spans five different dataset sizes to assess performance under varying conditions and computational constraints.

\subsection{Evaluation Framework
}
\label{subsec:evaluation_framework}

 We assess all approaches using three primary metrics: cache hit ratio, token savings ratio, and response time. This consistent evaluation framework enables direct comparison between our ensemble method, individual embedding models, and baseline fusion techniques such as averaging and concatenation.

\subsubsection{Initial Base Model Pool}
Our initial pool consists of five diverse embedding models selected to cover a broad spectrum of semantic representation capabilities while balancing performance and computational efficiency:

\begin{itemize}
    \item \textbf{all-MiniLM-L6-v2:} A lightweight 6-layer MiniLM model optimized for general semantic similarity tasks, offering fast inference and compact 384-dimensional embeddings suitable for resource-constrained environments.

    \item \textbf{all-MiniLM-L12-v2:} A deeper 12-layer variant of MiniLM, which enhances semantic understanding with increased model depth while maintaining efficiency, producing 384-dimensional embeddings with improved expressiveness over the 6-layer model.

    \item \textbf{paraphrase-MiniLM-L6-v2:} Fine-tuned extensively on paraphrase detection using over a billion sentence pairs, this model excels at identifying semantically equivalent but lexically diverse expressions, making it highly suitable for duplicate query detection.

    \item \textbf{all-mpnet-base-v2:} Based on the MPNet architecture \citep{song2020mpnet}, this model produces richer 768-dimensional embeddings by leveraging bidirectional attention to capture context more comprehensively, leading to strong performance across a wide range of NLP tasks.

    \item \textbf{paraphrase-distilroberta-base-v1:} A distilled version of RoBERTa \citep{liu2019roberta} fine-tuned for paraphrase identification, combining RoBERTa’s robust language understanding with the efficiency gains from distillation, generating 768-dimensional embeddings effective in semantic similarity applications.
\end{itemize}

To select base embedding models we employ a systematic approach to identify models that provide complementary semantic representations while maintaining computational efficiency.

\textbf{Correlation Analysis:} To understand how different embedding models relate to each other, we conduct a systematic correlation study. We start by selecting a balanced sample of 100,000 query pairs from the QQP dataset, ensuring we have an equal mix of similar and dissimilar question pairs to avoid bias in our analysis.
For our analysis, we process each query pair \(Q1, Q2\) through all the embedding models we want to compare. Each model transforms the queries into high-dimensional vectors and we calculate the cosine similarity between these embeddings, giving us a similarity score that indicates how closely related the two queries are according to that particular model.

This process generates what we call a "similarity score vector" for every query pair. Think of this vector as a fingerprint where each position corresponds to a different embedding model's assessment of how similar the two queries are. For example, if we're comparing 5 different embedding models, each query pair will have a 5-element vector containing the similarity scores from each model.

The final step involves measuring how consistently these models behave relative to each other. We compute pairwise correlations between all the embedding models' similarity scores across our entire dataset. This correlation analysis reveals which models tend to agree with each other and which ones provide distinctly different perspectives on query similarity. The results are visualized as a 5×5 heatmap, where each cell shows the correlation coefficient between two specific embedding models, helping us understand the landscape of model relationships and identify clusters of similar-behaving models \citep{bryan2024comparing, liu2024academic}.

\subsubsection{Base Model Selection}
\label{subsubsec:base_model_selection_setup}

\begin{figure}[h]
\centering
\includegraphics[width=\linewidth]{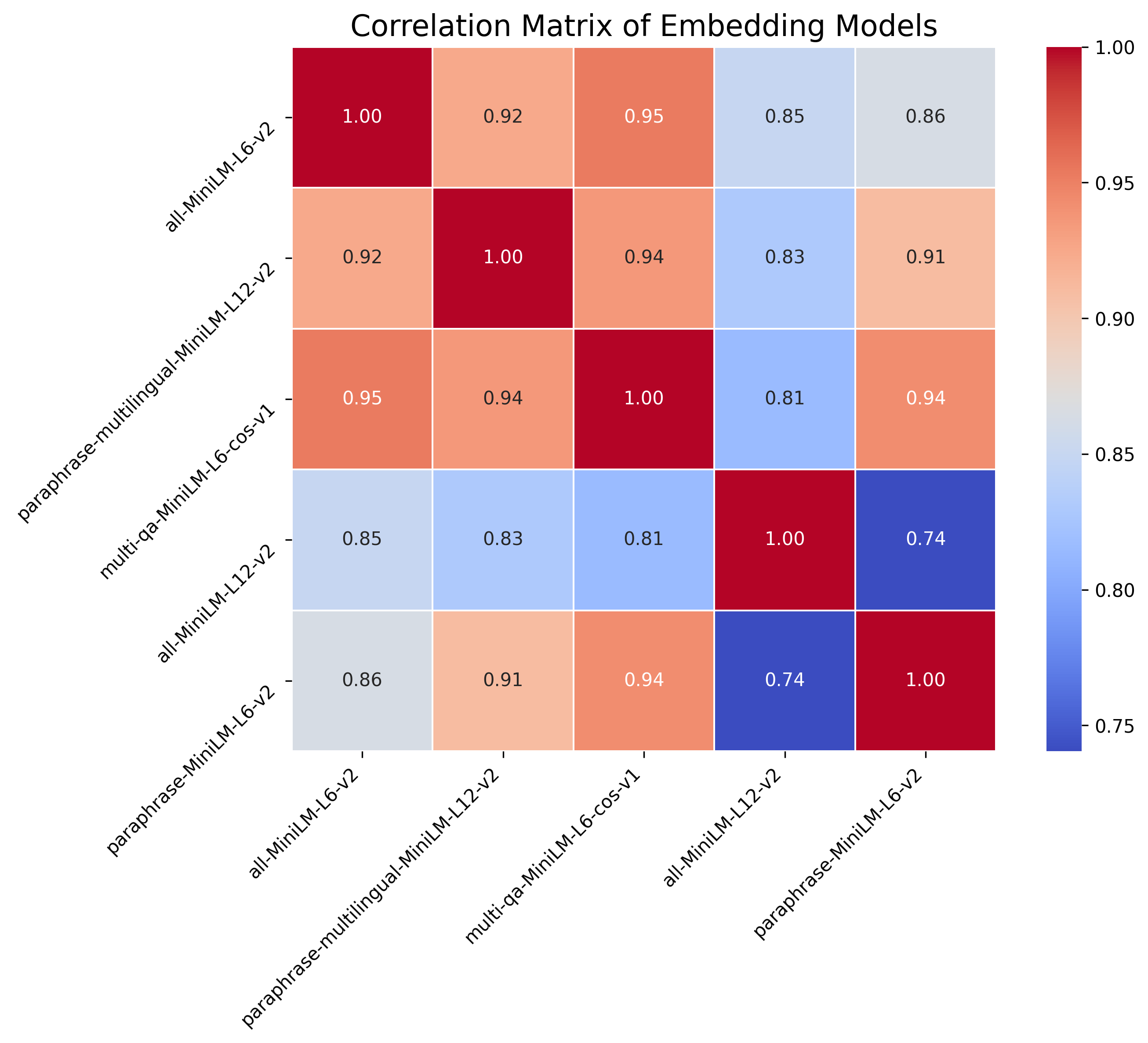}
\caption{Correlation matrix of base embedding models showing pairwise cosine similarity correlations across the evaluated models.}
\label{fig:correlation_matrix}
\end{figure}

Based on our correlation analysis, we select two embedding models that exhibit the lowest pairwise correlation (0.74) while maintaining strong individual performance. The selection of two models represents a practical balance between representational diversity and computational overhead. While additional models could further enhance semantic coverage, they would proportionally increase processing time and resource requirements for each query. As shown in Figure~\ref{fig:correlation_matrix}, these two models exhibit the lowest pairwise correlation among the evaluated candidates, reinforcing their complementary semantic representation capabilities.

\subsubsection{Baseline Comparisons}
\label{subsec:baseline_comparisons}

Our experimental evaluation includes several baseline approaches to provide comprehensive performance context:

\begin{itemize}
    \item \textbf{Individual Models:} Performance of each base embedding model operating independently, establishing single-model baselines.

    \item \textbf{Non-parametric Fusion:} Simple combination techniques including:
    \begin{itemize}
        \item \textbf{Averaging:} Element-wise averaging of embeddings from base models.
        \item \textbf{Concatenation:} Direct concatenation of base model embeddings.
    \end{itemize}

    \item \textbf{Ensemble Meta-Encoder:} Our proposed trainable approach that learns optimal combination strategies through supervised training.
\end{itemize}

\subsection{Main Results}
\label{subsec:main_results}

In this section, we validate the effectiveness of our proposed ensemble embedding encoder through a multi-stage analysis. We first evaluate its classification performance using accuracy, precision, recall, and F1-score for both duplicate and non-duplicate classes. Next, we assess its hit and miss ratios relative to selected individual embedding models. Finally, we compare it against baseline fusion methods—including averaging and concatenation—demonstrating that our trainable encoder consistently outperforms both individual models and non-trainable combinations.

\subsubsection{Encoder Performance}
\label{subsubsec:encoder_performance_vs_baselines}
Table~\ref{tab:classification_report} presents the detailed performance breakdown for both classes, demonstrating balanced performance across precision, recall, and F1-score metrics.

The results show that our encoder achieves an overall accuracy of 86\% with consistent performance across both duplicate (class 1.0) and non-duplicate (class 0.0) query pairs. The macro average metrics—calculated by treating both classes equally regardless of support—yield precision, recall, and F1-scores of 0.86 each. Since the dataset is perfectly balanced with 7,500 samples per class, the weighted averages are identical to the macro averages at 0.86 across all metrics, confirming no class imbalance effects.

Notably, the model exhibits complementary behavior across classes: non-duplicate queries (class 0.0) achieve higher precision (0.88) but lower recall (0.83), while duplicate queries (class 1.0) show the inverse pattern with 0.84 precision and 0.88 recall. This balanced trade-off indicates that the model does not exhibit significant bias toward either class, which is crucial for semantic caching applications where both false positives (incorrectly identifying non-duplicates as duplicates) and false negatives (missing actual duplicates) can negatively impact system performance and cache efficiency.

\setlength{\tabcolsep}{4pt} 
\begin{table}[h]
\centering
\begin{tabular}{|c|c|c|c|c|}
\hline
\textbf{Class} & \textbf{Precision} & \textbf{Recall} & \textbf{F1-Score} & \textbf{Support} \\
\hline
0.0 & 0.88 & 0.83 & 0.85 & 7500 \\
1.0 & 0.84 & 0.88 & 0.86 & 7500 \\
\hline
\textbf{Accuracy} & \multicolumn{4}{c|}{0.86 (Total samples: 15000)} \\
\textbf{Macro Avg} & 0.86 & 0.86 & 0.86 & 15000 \\
\textbf{Weighted Avg} & 0.86 & 0.86 & 0.86 & 15000 \\
\hline
\end{tabular}
\caption{Classification Report}
\label{tab:classification_report}
\end{table}

\subsubsection{Comparative Analysis}
\label{subsubsec:comparative_analysis}

To assess the effectiveness of our proposed ensemble meta-encoder, we compare it against widely used non-trainable fusion strategies that serve as baseline approaches. Below, we describe the fusion strategies evaluated in our analysis:

\textbf{Averaging:} Simple element-wise averaging of the \( n \) embeddings provides equal weight to all base models:
\[
e_{\text{avg}} = \frac{1}{n} \sum_{i=1}^n e_i
\]

\textbf{Concatenation:} Preserves all input information by sequentially appending the \( n \) embeddings, resulting in an \( n \times d \)-dimensional representation:
\[
e_{\text{concat}} = [e_1; e_2; \ldots; e_n]
\]

\subsubsection{Cache Hit and Miss Ratio Performance}
\label{subsubsec:hit_miss_ratio}

\begin{figure}[t]
\centering
\includegraphics[width=\linewidth]{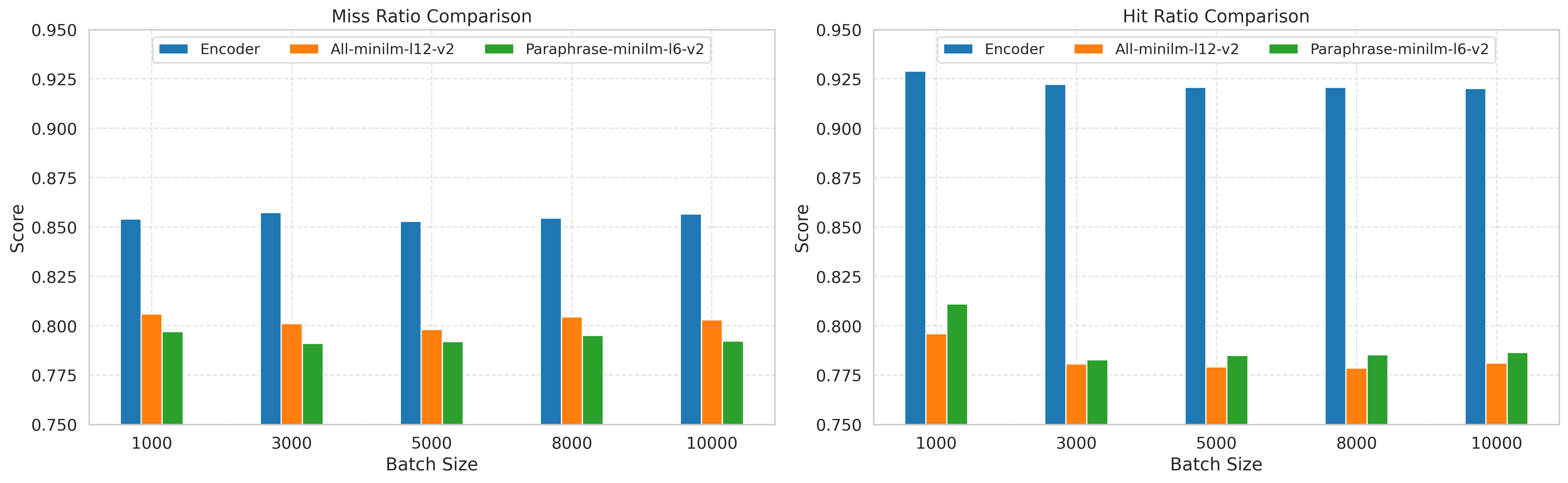}
\caption{Performance comparison of the proposed trainable encoder against individual embedding models.}
\label{fig:encoder_vs_individuall}
\end{figure}

We evaluate our trainable encoder against individual models and fusion baselines, focusing on cache hit and miss ratios across varying dataset sizes. Figure~\ref{fig:encoder_vs_individuall} shows that \textbf{paraphrase-MiniLM-L6-v2} achieves the highest hit ratio among individual models, while \textbf{all-MiniLM-L12-v2} performs better on miss ratios. However, our encoder consistently outperforms both in all cases.

\textbf{Relative Improvement:}
Table~\ref{tab:relative_improvements} summarizes the encoder’s relative improvement over the strongest individual model at each scale. On average, it achieves a 10.3\% boost in hit ratio and a 7.5\% gain in miss ratio.

\begin{table}[h!]
\centering
\caption{Relative improvement analysis of our trainable encoder across different dataset sizes.}
\label{tab:relative_improvements}
\begin{tabular}{|c|c|c|c|c|}
\hline
\textbf{Dataset Size} & \multicolumn{2}{c|}{\textbf{Hit Ratio}} & \multicolumn{2}{c|}{\textbf{Miss Ratio}} \\
\cline{2-5}
 & Encoder & Rel. Imp. & Encoder & Rel. Imp. \\
\hline
20\%  & 92.0\% & +8.9\% & 85.2\% & +6.4\% \\
40\%  & 91.8\% & +10.3\% & 85.0\% & +7.7\% \\
60\%  & 92.1\% & +11.2\% & 84.8\% & +9.8\% \\
80\%  & 92.3\% & +12.4\% & 85.3\% & +7.3\% \\
100\% & 92.0\% & +8.7\% & 85.1\% & +6.1\% \\
\hline
\textbf{Average} & \textbf{92.0\%} & \textbf{+10.3\%} & \textbf{85.1\%} & \textbf{+7.5\%} \\
\hline
\end{tabular}
\end{table}

\begin{figure}[t]
\centering
\includegraphics[width=\linewidth]{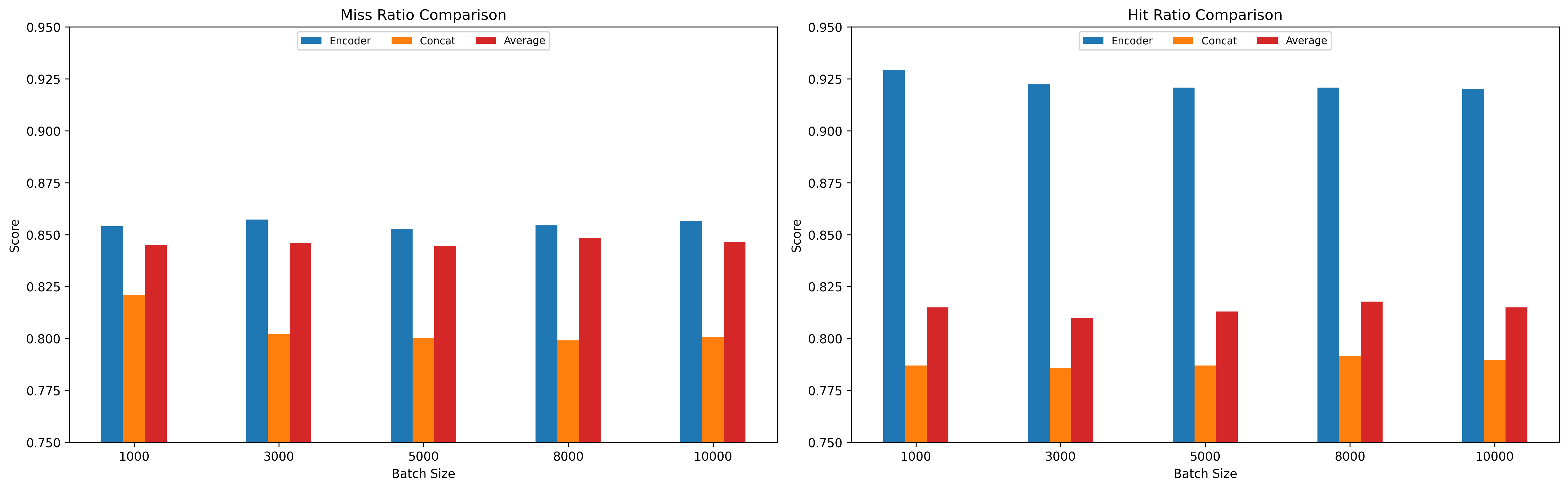}
\caption{Detailed comparison of our encoder against fusion strategies.}
\label{fig:encoder_vs_base_fusion}
\end{figure}

\textbf{Comparison with Fusion Methods:}
Figure~\ref{fig:encoder_vs_base_fusion} highlights that averaging offers modest gains, while concatenation underperforms due to dimensional inefficiency. Our encoder surpasses both: +7.2\% vs. averaging and +15.6\% vs. concatenation in hit ratio, with similar margins in miss ratio.

These results affirm that our ensemble encoder effectively combines diverse semantic signals, yielding robust and scalable performance across dataset sizes.

\subsubsection{Response Time}
\label{subsubsec:response_time}

\begin{figure}[h]
\centering
\includegraphics[width=\linewidth]{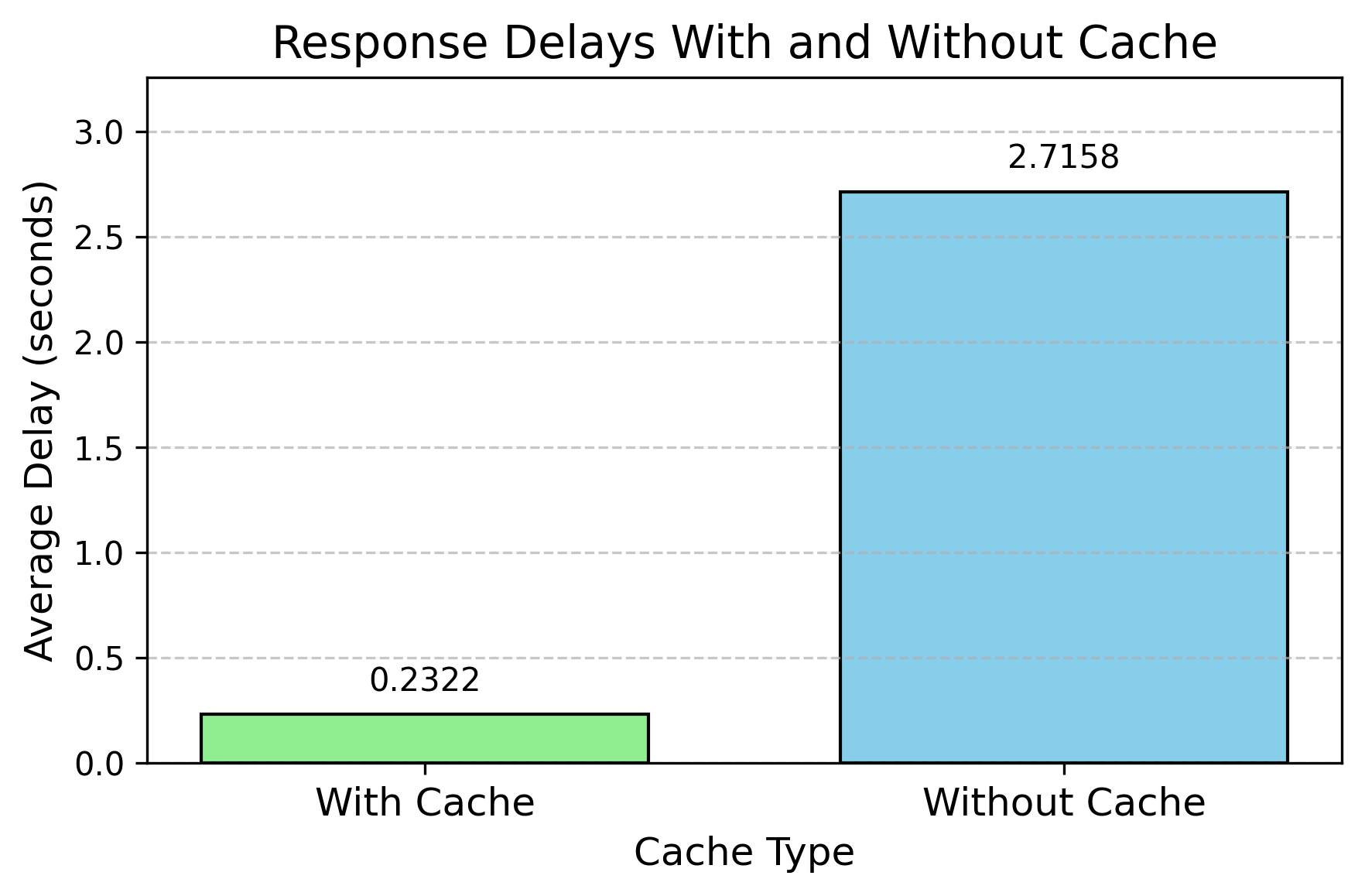}
\caption{Comparison of response times, demonstrating a reduction from 2.7 seconds to 0.3 seconds using semantic caching with the encoder-based ensemble method.}
\label{fig:response_time}
\end{figure}

As shown in Figure~\ref{fig:response_time}, the response time reduction from 2.7 seconds to 0.3 seconds was measured by randomly selecting 1,000 paired queries from duplicate samples, populating the cache with the first questions, and then requesting the 1,000 corresponding questions. The average response time of 0.3 seconds represents how quickly the cached responses were served using the proposed ensemble method, compared to the baseline scenario where all 1,000 questions were served directly by the LLM at an average of 2.7 seconds per request. This demonstrates the efficiency gain when semantically similar queries are successfully identified and served from cache rather than requiring full LLM inference.

\subsubsection{Token Savings}
\label{subsubsec:token_savings}

The system achieves approximately 20\% token savings, calculated using the same 1,000 duplicate paired questions with the populated cache system. This percentage represents the average proportion of total tokens that were served by the cache instead of being generated by the LLM. When queries are successfully identified as duplicates and served from cache, both input tokens (query processing) and output tokens (response generation) are completely avoided.

\subsubsection{Eviction Policy Impact}
\label{subsubsec:eviction_policy_impact}

When the cache reaches capacity, eviction is triggered based on the selected policy \citep{taiye2024algorithmic, callon2024practice}.

\begin{figure}[h]
\centering
\includegraphics[width=\linewidth]{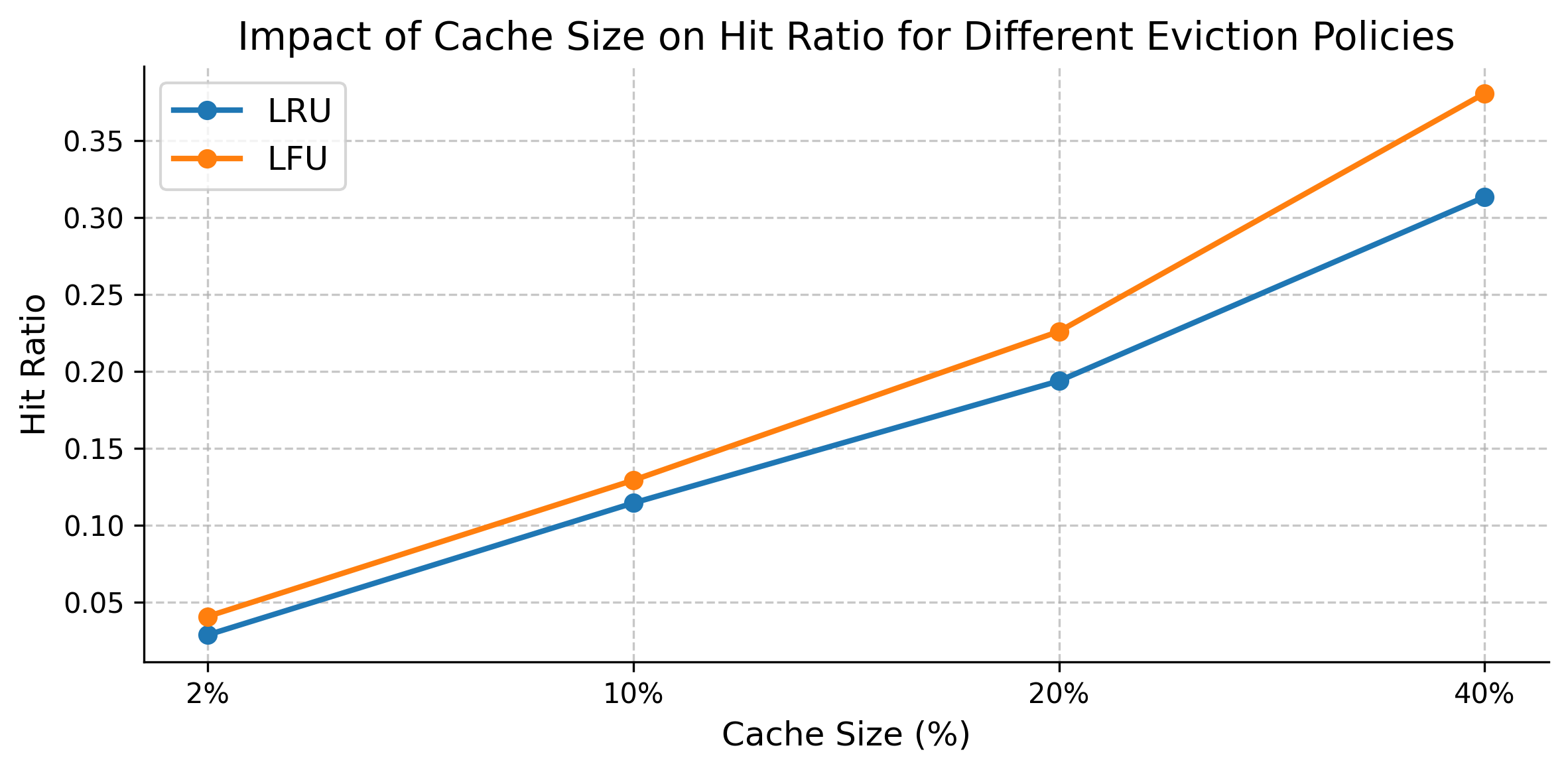}
\caption{Hit ratio versus cache size (percentage of total queries) for Least Recently Used (LRU) and Least Frequently Used (LFU) eviction policies. LFU typically achieves higher hit ratios across cache sizes.}
\label{fig:eviction_policy_impact}
\end{figure}

Figure~\ref{fig:eviction_policy_impact} illustrates the impact of cache size—expressed as a percentage of 10,000 total queries—on the hit ratio for LRU and LFU algorithms \citep{belady1966study}. LFU generally outperforms LRU across different query scales, though the optimal choice depends on request patterns and embedding quality.

\section{Conclusion and Future Work}
\label{sec:conclusion}

This paper presents a novel ensemble embedding approach for semantic caching in LLM-based systems that combines multiple embedding models through a trainable meta-encoder. Our method achieves a 92\% cache hit ratio for duplicate queries and 85\% accuracy for non-duplicate queries, representing substantial improvements over single-model approaches with an average 10.3\% boost in hit ratio and 7.5\% gain in miss ratio.
Compared to traditional fusion methods, our approach outperforms averaging by 7.2\% and concatenation by 15.6\% in hit ratio. The practical impact includes dramatic response time reduction from 2.7 seconds to 0.3 seconds and approximately 20\% token savings. These consistent improvements across varying dataset sizes demonstrate the effectiveness of ensemble embeddings in enhancing caching performance and reducing computational costs in LLM-based systems.

For future work, we plan to implement a more systematic query eviction policy for when the cache reaches its capacity, aiming to outperform traditional methods such as LRU and LFU \citep{belady1966study} in hit ratio. Currently, our implementation does not include privacy-preserving measures \citep{wang2024meancache}, which could compromise users' query safety. To address this limitation, our future developments will focus on ensuring that users' data remains secure \citep{robbins2024invention, chen2024topical}.

\bibliographystyle{elsarticle-num}

\bibliography{manuscript}
\end{document}